\documentclass{article}
\usepackage[square,numbers,compress]{natbib}
\PassOptionsToPackage{numbers, compress}{natbib}

\usepackage[preprint]{neurips_2024}

\usepackage{longtable}

\usepackage[utf8]{inputenc} 
\usepackage[T1]{fontenc}    
\usepackage{hyperref}       
\usepackage{url}            
\usepackage{booktabs}       
\usepackage{amsfonts}       
\usepackage{nicefrac}       
\usepackage{microtype}      
\usepackage{xcolor}         

\usepackage{url}
\usepackage{subfigure}
\usepackage{multirow}
\usepackage{enumitem}
\usepackage{stmaryrd}
\usepackage{array}
\usepackage{threeparttable}
\usepackage{blindtext}
\usepackage{amssymb}

\usepackage[ruled,vlined,linesnumbered,longend]{algorithm2e}
\usepackage{svg}
\usepackage{placeins}

\usepackage{soul}
\usepackage[utf8]{inputenc}
\usepackage{graphicx}
\usepackage{amsmath}
\usepackage{booktabs}
\usepackage{lipsum}
\usepackage{listings}
\usepackage{xcolor, soul}
\sethlcolor{blue!10}
\usepackage{colortbl}

\usepackage{wrapfig,lipsum,booktabs}
\usepackage{changepage}  

\newcommand{\ourmethod}{\textsc{ARTE}\xspace}

\usepackage{amsmath,amsfonts,bm}

\def\eqref#1{equation~\ref{#1}}

\def\1{\bm{1}}

\DeclareMathAlphabet{\mathsfit}{\encodingdefault}{\sfdefault}{m}{sl}
\SetMathAlphabet{\mathsfit}{bold}{\encodingdefault}{\sfdefault}{bx}{n}

\title{
Aligning Teacher with Student Preferences\\ for Tailored Training Data Generation
}

\author{%
  Yantao Liu\textsuperscript{*}$^\dagger$,\;\;Zhao Zhang\textsuperscript{*}$^\dagger$,\;\;Zijun Yao\textsuperscript{*},\;\;Shulin Cao,\;\;Lei Hou,\;\;Juanzi Li \\
Department of Computer Science and Technology, BNRist;\\
KIRC, Institute for Artificial Intelligence,\\
Tsinghua University, Beijing, 100084, China \\
  \texttt{yaozj20@mails.tsinghua.edu.cn, \{houlei,lijuanzi\}@tsinghua.edu.cn} \\
}

\begin{document}
\renewcommand{\thefootnote}{\fnsymbol{footnote}}
\footnotetext[1]{Equal contribution.}
\footnotetext[2]{Work was done during their internship at Tsinghua University.}

\maketitle

\begin{abstract}
Large Language Models (LLMs) have shown significant promise as copilots in various tasks. 
Local deployment of LLMs on edge devices is necessary when handling privacy-sensitive data or latency-sensitive tasks. 
The computational constraints of such devices make direct deployment of powerful large-scale LLMs impractical, necessitating the Knowledge Distillation from large-scale models to lightweight models.
Lots of work has been done to elicit diversity and quality training examples from LLMs, but little attention has been paid to aligning teacher instructional content based on student preferences, akin to ``responsive teaching'' in pedagogy. 
In response, we propose \textbf{\ourmethod}, dubbed \textbf{A}ligning Teache\textbf{R} with Studen\textbf{T} Preferenc\textbf{E}s, a framework that \textit{aligns the teacher model with student preferences to generate tailored training examples for Knowledge Distillation}.
Specifically, we first elicit draft questions and rationales from the teacher model, then collect student preferences on these questions and rationales using students' performance with in-context learning as a proxy,
and finally align the teacher model with student preferences with Direct Preference Optimization.
In the end, we repeat the first step with the aligned teacher model to elicit tailored training examples for the student model on the target task.
Through extensive experiments on academic benchmarks, we demonstrate the superiority of our method over existing instruction-tuning datasets distilled from powerful LLMs. 
Moreover, we thoroughly investigate the generalization of \ourmethod, including the generalization of fine-tuned student models in reasoning ability and the generalization of aligned teacher models to generate tailored training data across tasks and students.
In summary, our contributions lie in proposing a novel framework for tailored training example generation, demonstrating its efficacy in extensive experiments, and investigating the generalization of both fine-tuned student and aligned teacher models in \ourmethod.
The code is released at \url{https://github.com/THU-KEG/ARTE}.
\end{abstract}

\section{Introduction}
\label{sec:intro}

Recently, Large Language Models (LLMs) have demonstrated significant potential as copilots for various tasks~\cite{achiam2023gpt4,xie2023openagents}. 
Some applications involve sensitive data, such as medical diagnostics or legal advice~\cite{xie2023openagents,qian2023communicative}, or require low latency, such as robot control or autonomous driving~\cite{brohan2023rt,driess2023palme}. 
This necessitates the deployment of LLMs on computationally constrained edge devices, like smartphones or laptops. 
However, most existing successes~\cite{wang2023describe,qian2023communicative, park2023generative} depend on powerful, large-scale LLMs~\cite{achiam2023gpt4,chowdhery2023palm,touvron2023llama2}, which are computationally prohibitive for edge deployment. 
This scenario highlights the need for \textit{Knowledge Distillation}~\cite{hinton2015distilling}, a method to distill specific capabilities such as reasoning, planning, and decision-making from large-scale LLMs to lightweight models suitable for edge deployment.

There are two steps in Knowledge Distillation in LLMs: 1) Knowledge Elicitation, where the training data is curated from the teacher model
, and 2) Supervised Fine-Tuning, where the student model learns from the generated examples.
Since the research paradigm in the era of LLMs has shifted from model-centric to data-centric~\cite{zha2023data}, the key challenge lies in step 1 curation of training data.
\citet{hsieh-etal-2023-step-by-step,mukherjee2023orca} proposed to guide the teacher language model to generate a reasoning process with chain-of-thought~\cite{wei2022chain,yao2022react}, which provides more information for the student model to imitate.
\citet{wang2022self} utilize the ROUGE-L similarity~\cite{lin-2004-rouge} to filter out the redundant examples to ensure the diversity.
Inspired by the high quality of the textbook, \citet{li2023phi1.5} proposed to synthesize ``textbook-like'' text from the teacher model.
\citet{mitra2023orca2} contends that different tasks should be combined with different reasoning strategies when generating rationales of examples.
While these works have advanced in curating training examples, 
they overlook aligning the teacher model with responses from the student model to refine the training data,
referred to as ``responsive teaching'' in pedagogy~\cite{gay2000culturally}, which effectively lowers the learning barriers for students~\cite{hattie2007power}.

\begin{figure}[t]
    \begin{adjustwidth}{-0.05\textwidth}{-0.05\textwidth}  
        \centering
        \includegraphics[width=1.0\textwidth]{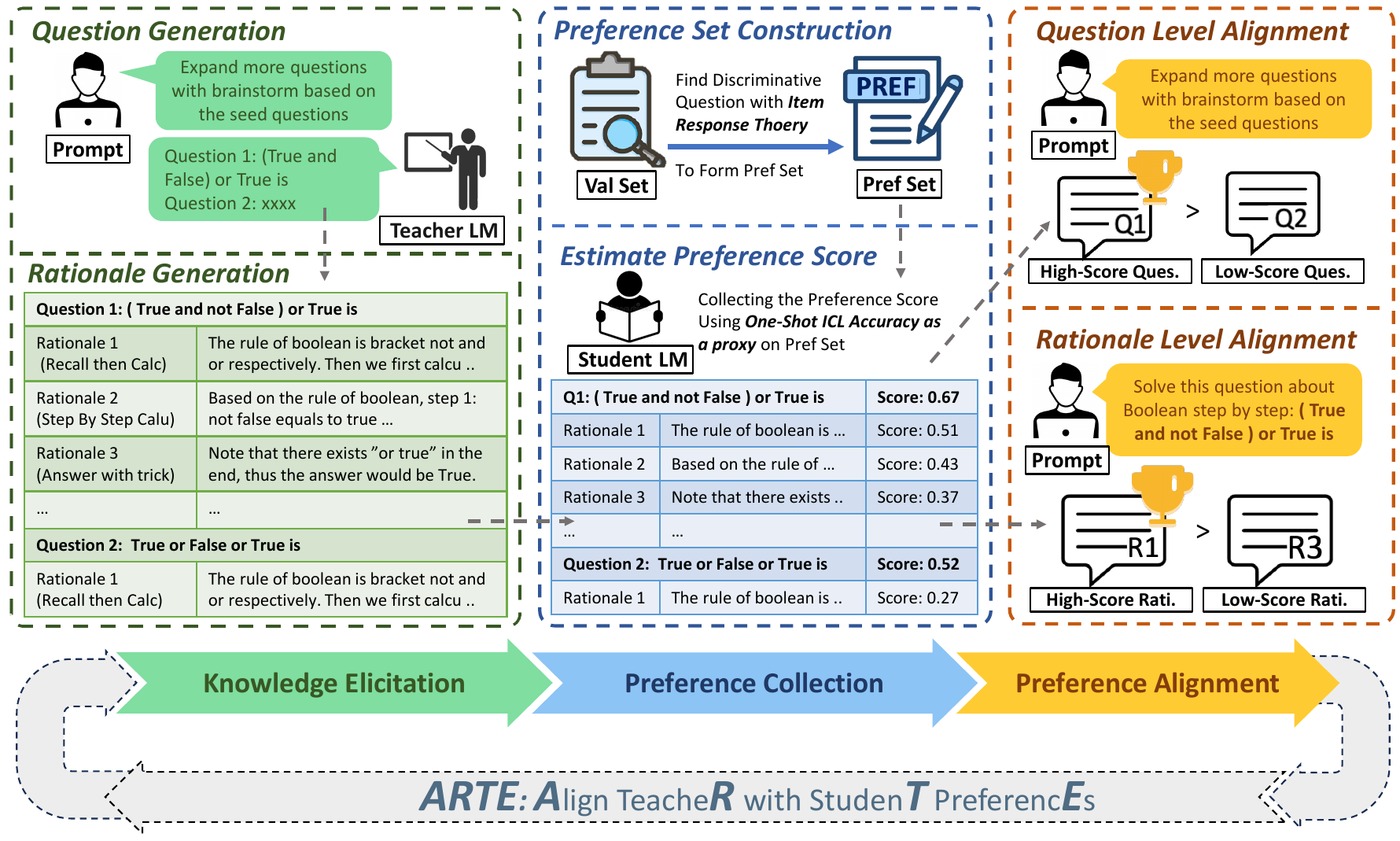}
        \vspace*{-0.3cm}
        \caption{The overall framework of \ourmethod.}
        \vspace*{-0.3cm}
        \label{fig:framework}
    \end{adjustwidth}
\end{figure}

To this end, we propose \textbf{\ourmethod},  dubbed \textbf{A}ligning Teache\textbf{R} with Studen\textbf{T} Preferenc\textbf{E}s, a novel framework that \textit{aligns the teacher language model with the student language model's preferences to generate tailored training examples for Knowledge Distillation}.
Our framework is illustrated in Figure~\ref{fig:framework}.
There are three main steps in our framework: 
1) \textit{Knowledge Elicitation}~(Section \ref{sec:generation}): we prompt the teacher model with seed questions to generate a set of draft examples consisting of question-rational pairs.
2) \textit{Preference Collection}~(Section \ref{sec:reward}): we collect the preference of each draft example through one-shot in-context learning on the preference set which contains the top-$k$ most discriminative questions from the validation set of the target task.
3) \textit{Preference Alignment}~(Section \ref{sec:alignment}): we align the teacher model with the student model's preferences through Direct Preference Optimization (DPO)~\cite{rafailov2023dpo} to improve instructional content.
Finally, we repeat the first step with the aligned teacher model to curate tailored training examples for the student model and use them to Supervised Fine-Tune the student model.

Extensive experiments in academic reasoning benchmarks show that our method outperforms existing instruction-tuning datasets~\cite{peng2023instruction,ivison2023camels,mukherjee2023orca,xu2023wizardlm} by a large margin.
Specifically, compared to the state-of-the-art instruction-tuning dataset, our method achieves improvements of $9.6\%$, $1.0\%$, $0.8\%$ and $8.5\%$ on the logic reasoning, commonsense reasoning, math reasoning, and knowledge reasoning tasks respectively in Big-Bench-Hard~\cite{suzgun-etal-2023-challenging} benchmark.
We also testify the generalization of \ourmethod in two aspects:
1) We investigate generalization in reasoning of fine-tuned student models with in-context learning in out-of-domain reasoning benchmarks, including PIQA~\cite{bisk2020piqa}, CommonsenseQA~\cite{talmor-etal-2019-commonsenseqa}, ARC-Easy~\cite{clark2018think}, ARC-Challenge~\cite{clark2018think}, and GSM8K~\cite{cobbe2021gsm8k}.
Results show that \ourmethod still outperforms baselines, testifying the effectiveness of \ourmethod in enhancing the reasoning ability of the student model.
2) We further investigate the generalization of the aligned teacher model across tasks and students.
The result provides two insights: First, the teacher model aligned in BBH can generate tailored training examples on unseen reasoning benchmarks, such as ARC or GSM8K.
Second, the teacher model aligned with Gemma-2B can generate tailored training examples for different student models from different families or different domains as long as they share a similar parameter capacity, such as Qwen1.5-1.8B or CodeGemma-2B.

To summarize, our contributions are three-fold:
1) Inspired by the responsive teaching in pedagogy, we propose \ourmethod, a novel framework that aligns the teacher language model with the student language model's preferences to generate tailored training examples for Knowledge Distillation.
2) Extensive experiments in in-domain and out-of-domain reasoning benchmarks show the student model fine-tuned with tailored training examples generated by \ourmethod outperforms existing instruction-tuning datasets by a large margin.
3) We also investigate the generalization of the aligned teacher model across tasks and students.
The results show that the aligned teacher model can curate tailored training examples across different reasoning tasks and different student models with similar parameter capacities.

\section{Preliminaries}
\paragraph{Knowledge Distillation}
Knowledge distillation is a widely used technique to transfer knowledge from a large teacher model to a smaller student model~\cite{hinton2015distilling}.
In this study, we focus on scenarios where both teacher and student models are pre-trained autoregressive language models. 
Given a teacher language model $LM_t$, a student language model $LM_s$, and a target task $T$, 
the goal of knowledge distillation is to improve the performance of the student model $LM_s$ on task $T$ by fine-tuning it with the training examples generated by the teacher model $LM_t$.
There are only a few labeled examples $D$ served as validation set available (e.g., 100 examples) for the target task $T$, each example consists of a question and an answer pair $(q, a)$.

\vspace*{-0.18cm}
\paragraph{Alignment of LLMs}
Alignment of LLMs aims to align the language model with human or AI preferences.
In this study, we focus on the scenario where the teacher model $LM_t$ is aligned with the student model $LM_s$'s preferences to generate tailored training examples for Knowledge Distillation.
Direct Preference Optimization (DPO)~\cite{rafailov2023dpo} is selected to align the teacher model with the student model's preferences due to its stability and simplicity compared to other methods like PPO~\cite{schulman2017proximal}.
It can be formulated as a simple loss function that can be optimized directly on a dataset of preference pairs ${(x,y_w,y_l)}$, where $x$ is an input prompt and $y_w$ and $y_l$ are the preferred and dispreferred responses, respectively.
DPO trains the teacher model to align with the student model's preferences by minimizing the loss function.

\section{\ourmethod}
We introduce \ourmethod, dubbed \textbf{A}ligning Teache\textbf{R} with Studen\textbf{T} Preferenc\textbf{E}s, a novel framework that aligns the teacher $LM_t$ with student $LM_s$'s preferences to generate tailored training examples in the target task $T$.
Our overall framework is illustrated in Figure~\ref{fig:framework}.
There are three main steps in our framework: 
1) \textit{Knowledge Elicitation}: we prompt the teacher model with seed questions to generate a draft dataset consisting of draft questions and corresponding draft rationales.
2) \textit{Preference Collection}: We collect the preference scores from the student model for both draft questions and rationales using the one-shot in-context learning performance of the student model on the preference set as a proxy.
3) \textit{Preference Alignment}: Based on the preference score of questions and rationales, we align the teacher model with the student model's preferences through Direct Preference Optimization (DPO).
Finally, we repeat the first step with the aligned teacher model to curate tailored training examples for the student model and use them to Supervised Fine-Tune the student model.
In the following, we will elaborate on each step in detail.

\subsection{Knowledge Elicitation}
\label{sec:generation}
In this step, we construct a set of draft dataset $D_\text{draft} = \{(q, R_q)\}$ for the target task $T$, where $q$ is the draft question and $R_{q} = \{r_q\}$ is the set of rationales associated with the question $q_i$.
For question generation,
we prompt the teacher model $LM_t$ with seed questions to expand them into $m$ draft questions.
Specifically, we use three seed questions to construct one question generation prompt which guides the teacher model to brainstorm multiple questions.
We decode the question generation prompt at a temperature $1.0$ multiple times until we get $m$ draft questions.

For rationale generation, inspired by the observation that one question may have multiple different reasoning processes leading to the same answer and the optimal reasoning process may vary for different questions and different models~\cite{mitra2023orca2}.
Then, for each draft question $q$, $LM_t$ is guided to generate $n$ rationales $r_q$ to form a rationale set $R_q$ using different reasoning techniques, such as \textit{Explain Like I'm Five}, \textit{Step-by-Step}, or \textit{Math Symbols}.
Finally, we combine each draft question $q$ with its corresponding rationales $R_q$ to form a draft sample $(q, R_q)$.
Since each draft question $q$ would be expanded into $n$ rationales, there are $m \times n$ draft examples $E_\text{draft} = \{(q, r_q)\}$ in the draft dataset $D_\text{draft}$.

\subsection{Preference Collection}
\label{sec:reward}
In this step, our target is to collect the preference of the student model in both question and rationale levels.
Namely, we aim to determine which question or rationale is more likely to be accepted by the student model during the fine-tuning process. 
\citet{dai-etal-2023-gpt} points out that language models secretly perform gradient descent w.r.t the in-context examples during in-context learning like supervised fine-tuning.
Inspired by this, we contend that the student model's preference for one draft example $(q, r)$ in supervised fine-tuning can be reflected by the performance of the student model with $(q, r)$ as the example in one-shot in-context learning.

Therefore, we use the in-context performance of the student model with draft examples on the validation set $D_\text{val}$ as the preference score.
Specifically, for rationale $r$ whose corresponding question is $q$, the preference score $s_r$ is defined as follows:
\begin{equation}
    s_r = \frac{1}{|D_\text{val}|} \sum_{(q', a') \in D_\text{val}} \mathbb{I}\left(\text{LM}_s(q, r; q') = a'\right)
\end{equation}
where $\text{LM}_s(q, r; q')$ denotes the student model's answer to the question $q'$ when $q$ and $r$ are used as the example in one-shot in-context learning, and $\mathbb{I}(\cdot)$ is the indicator function.
Then the preference score $s_q$ of question $q$ is defined as the average of the preference scores of its associated rationales:
\begin{equation}
    s_q = \frac{1}{|R_q|} \sum_{r \in R_q} s_r
\end{equation}
where $R_q$ is the set of rationales associated with the question $q$.

However, the computation cost would be high if we evaluate the student model on the entire validation set for all $m \times n$ draft examples.
Existing alignment methods such as DPO~\cite{rafailov2023dpo} and PPO~\cite{schulman2017proximal} only require understanding the partial order of outputs relative to a single input to learn preferences. 
Inspired by this, we aim to sample a small, discriminative subset of questions from the validation set to estimate their preference scores effectively.
Specifically, Item Response Theory (IRT)~\cite{polo2024tinybenchmarks,embretson2013item} is adopted to measure the discriminability of each question in the validation set.
First, we collect a scoring matrix $S \in \mathbb{R}^{a \times b}$ from $a$ different LLMs on $b$ samples from the validation set.
Then, we use the two-parameter logistic model~\cite{thissen2001item} to estimate the discrimination parameter $\alpha$ and the difficulty parameter $\beta$ for each question based on the score matrix $S$ through bayesian inference~\cite{natesan2016bayesian,lalor2019emnlp,rodriguez2021evaluation}.
Finally, we select the top-$k$ most discriminative questions as the preference set $D_\text{pref}$.
Thus the estimated preference score $\hat{s}_r$ and $\hat{s}_q$ can be formulated as:
\begin{equation}
    \hat{s}_r = \frac{1}{|D_\text{pref}|} \sum_{(q', a') \in D_\text{pref}} \mathbb{I}\left(\text{LM}_s(q, r; q') = a'\right)
\end{equation}
and
\begin{equation}
    \hat{s}_q = \frac{1}{|R_q|} \sum_{r \in R_q} \hat{s}_r
\end{equation}
Without special note, we refer to $\hat{s}_r$ and $\hat{s}_q$ as reference scores in the following sections.

\subsection{Preference Alignment}
\label{sec:alignment}
After collecting the preference score of questions and rationales, we aim to align the teacher model with the student model's preferences to generate tailored examples for the target task $T$.
There are two main aspects to align, \textit{Question Generation} and \textit{Rationale Generation}, towards generating not only tailored rationales but also tailored questions. 
Direct Preference Optimization (DPO)~\cite{rafailov2023dpo} is selected to align the teacher model with the student model's preferences due to its stability and simplicity compared to other methods like PPO~\cite{schulman2017proximal}.
For question level alignment, for the input question generation prompt, we randomly sample out $k$ draft questions with the top-$25\%$ and bottom-$25\%$ preference scores as the chosen and rejected responses respectively.
For rationale level alignment, for each question $q$, we select the rationale from $R_q$ with the highest preference score as the chosen response and the rationale with the lowest preference score as the rejected response.
Note that the input prompt in rationale level alignment is just a naive step-by-step prompt as shown in Table~\ref{tab:dpo-rationale}.
Finally, we mix the question level and rationale level DPO datasets to perform DPO to align the teacher model with the student model's preferences.

After the alignment, we repeat the first step to generate tailored examples for the target task $T$ and use them to Supervised Fine-Tune the student model.
Note that when generating tailored questions and rationales, we use the same input prompt from the alignment step and decode at temperature $\tau=0$ for deterministic decoding. More details can be found in Appendix~\ref{sec:repeating}.


\section{Experiments}
\label{sec:experiments}
In this section, we conduct experiments to evaluate the effectiveness of our method to transfer the reasoning capabilities from the teacher model to the student model to target tasks Big-Bench-Hard~\cite{suzgun-etal-2023-challenging}.
\subsection{Experimental Setup}

\paragraph{Datasets}
We conduct experiments on the Big-Bench-Hard~\cite{suzgun-etal-2023-challenging}, a suite of 23 challenging tasks from the Big-Bench benchmark~\cite{srivastava2022beyond}. 
The tasks are designed to test the generalization ability of models on a wide range of tasks, including logical reasoning, commonsense reasoning, world knowledge, math ability, etc.
For better presentation, we divide the tasks into four categories based on the capabilities required by the tasks: (1) \textit{logical reasoning}, (2) \textit{commonsense reasoning}, (3) \textit{math reasoning} and (4) \textit{world knowledge}, denoted as \textit{BBH-Logic}, \textit{BBH-CS}, \textit{BBH-Math} and \textit{BBH-Knowl} respectively.
Details of the task categories are shown in Appendix~\ref{sec:bbh-category}.
Table~\ref{tab:bbh} shows the statistical information of the tasks in Big-Bench-Hard.
\begin{wraptable}{r}{0.50\textwidth}
    \centering
    \begin{tabular}{lccc}
        \toprule
        \textbf{Category} & \textbf{\#Tasks} & \textbf{\#Pref} & \textbf{\#Test} \\
        \midrule
        BBH-Logic & 10 & 400 &  3,146\\
        BBH-CS & 7 & 280 &  1,615\\
        BBH-Math & 3 & 120 &  750\\
        BBH-Knowl & 3 & 120 &  750\\
        \midrule
        Total & 23 & 920 &  6,261\\
        \bottomrule
    \end{tabular}
    \caption{Statistics of the Big-Bench-Hard dataset.}
    \vspace*{-0.3cm}
    \label{tab:bbh}
\end{wraptable}

As mentioned in Section~\ref{sec:reward}, we collect the top-$40$ most discriminative questions in each task from the validation set of the Big-Bench-Hard to form the preference set to reduce the computation cost.
Specifically, first, sampling $100$ questions for each task from original Big-Bench dataset to form the validation set.\footnote{For \textit{Causal Judgment}, \textit{Penguins in a Table}, and \textit{Snarks}, there is no extra samples apart from the test set in BBH, thus we use \texttt{GPT-4} to generate the validation set and manually filter out the questions with low quality.}
Second, we collect the detailed results of Gemma (2B,7B), CodeGemma-2B, Qwen1.5-1.8B, and MiniCPM-2B on the validation set under 0,1,2,3-shot settings.
Finally, we use the 2P-IRT model to find out the top-$40$ most discriminative questions to form the preference set for each task and testify the consistency of preference scores collected from the preference set and the original validation set. The details are in Appendix~\ref{sec:consistency}.

\paragraph{Baselines}
We compare our method with the following instruction-tuning datasets as academic baselines:
1) GPT-4-LLM~\cite{peng2023instruction} is an instruction-tuning dataset distilled from \texttt{GPT-4} with Self-Instruct~\cite{wang2022self}.
2) Tulu-v2~\cite{ivison2023camels} is a mix of multiple existing high-quality instruction-tuning datasets, including FLAN~\cite{longpre2023flan}, OpenAssistant~\cite{kopf2024openassistant} etc.
3) OpenOrca is a reproduction of Orca~\cite{mukherjee2023orca}, which augments FLAN data with additional \texttt{GPT-4} or \texttt{GPT-3.5-Turbo} generated explanations.
4) WizardLM-Evol-Instruct~\cite{xu2023wizardlm} is instruction-fine-tuning dataset which is distilled from the \texttt{GPT-4} through \textit{Evol-Instruct}.

Apart from the above academic baselines, we also compare our method with ablation baselines to testify the effectiveness of the two different alignment aspects in our method:
1) \textit{Orignal Teacher}: the dataset constructed by the draft questions and its corresponding rationale with the highest preference score in the preference set.
Namely, in this baseline, both the questions and rationales are generated by the original teacher model.
2) \textit{Rationale Only}: In this setting, only the rationales are generated by the aligned teacher model, while the questions are generated by the original teacher model.
3) \textit{Question Only}: In this setting, only the questions are generated by the aligned teacher model, while the rationales are generated by the original teacher model.
To ensure a fair comparison, we sample out $6,750$ examples from each baseline dataset to conduct the experiment.
Supervised Fine-Tune is used to train the student model on these datasets with 1 epoch using the Adam optimizer~\cite{kingma2014adam} and a learning rate of $2e-5$.
We use 3\% of the total training steps for cosine learning rate warm-up and adopt a linear decay learning rate schedule.

\begin{table}[t]
\centering
\label{tab:bbh-results}
\scalebox{0.95}{
\begin{tabular}{lcccc|c}
\toprule
\textbf{Models}   & \textbf{BBH-Logic} & \textbf{BBH-CS} & \textbf{BBH-Math} & \textbf{BBH-Knowl} & \textbf{BBH} \\ 
\midrule
\multicolumn{6}{c}{\textit{zero-shot result}}                                      \\
\midrule
Vanilla Gemma-2B & 0.80      & 0.00            & 6.00     & 0.00          & 1.09\\
\midrule
~+ GPT-4-LLM & \cellcolor[HTML]{deebf7}1.35 & 0.00 & \cellcolor[HTML]{FDB999}1.60 & \cellcolor[HTML]{deebf7}0.93 & \cellcolor[HTML]{FDE9D9}1.00 \\
~+ Tulu-v2 & \cellcolor[HTML]{c6dbef}6.15 & \cellcolor[HTML]{c6dbef}5.83 & \cellcolor[HTML]{9ecae1}16.00 & \cellcolor[HTML]{deebf7}0.67 & \cellcolor[HTML]{c6dbef}6.33 \\
~+ WizardLM & \cellcolor[HTML]{c6dbef}6.24 & \cellcolor[HTML]{deebf7}2.14 & \cellcolor[HTML]{9ecae1}12.27 & \cellcolor[HTML]{9ecae1}15.33 & \cellcolor[HTML]{c6dbef}6.88 \\
~+ OpenOrca & \cellcolor[HTML]{c6dbef}5.17 & \cellcolor[HTML]{c6dbef}4.68 & \cellcolor[HTML]{9ecae1}20.13 & \cellcolor[HTML]{c6dbef}8.53 & \cellcolor[HTML]{c6dbef}7.42 \\
\midrule
~+ Orignal Teacher & \cellcolor[HTML]{9ecae1}35.57 & \cellcolor[HTML]{9ecae1}40.61 & \cellcolor[HTML]{9ecae1}33.87 & \cellcolor[HTML]{9ecae1}40.27 & \cellcolor[HTML]{9ecae1}37.55 \\
~+ Rationale Only & \cellcolor[HTML]{9ecae1}36.07 & \cellcolor[HTML]{9ecae1}37.54 & \cellcolor[HTML]{9ecae1}37.87 & \cellcolor[HTML]{9ecae1}40.53 & \cellcolor[HTML]{9ecae1}37.94 \\
~+ Question Only & \cellcolor[HTML]{6baed6}\textbf{40.29} & \cellcolor[HTML]{9ecae1}43.45 & \cellcolor[HTML]{9ecae1}40.80 & \cellcolor[HTML]{9ecae1}40.67 & \cellcolor[HTML]{9ecae1}41.73 \\
\midrule
~+ \ourmethod & \cellcolor[HTML]{6baed6}39.07 & \cellcolor[HTML]{6baed6}\textbf{44.44} & \cellcolor[HTML]{6baed6}\textbf{42.00} & \cellcolor[HTML]{6baed6}\textbf{42.53} & \cellcolor[HTML]{6baed6}\textbf{41.96} \\
\midrule
\multicolumn{6}{c}{\textit{three-shot result}}                                      \\
\midrule
Vanilla Gemma-2B & 28.16 & 40.92 & 42.13 & 44.40 & 35.57 \\
\midrule
~+ GPT-4-LLM & \cellcolor[HTML]{9ecae1}29.58 & \cellcolor[HTML]{FDE9D9}40.79 & \cellcolor[HTML]{FDB999}36.27 & \cellcolor[HTML]{9ecae1}49.20 &
\cellcolor[HTML]{9ecae1}36.11 \\
~+ Tulu-v2 & \cellcolor[HTML]{9ecae1}30.28 & \cellcolor[HTML]{FDE9D9}39.74 & \cellcolor[HTML]{FDE9D9}38.67 & \cellcolor[HTML]{9ecae1}50.00 & \cellcolor[HTML]{9ecae1}36.35 \\
~+ WizardLM & \cellcolor[HTML]{9ecae1}29.62 & \cellcolor[HTML]{9ecae1}43.44 & \cellcolor[HTML]{9ecae1}45.33 & \cellcolor[HTML]{9ecae1}45.33 & \cellcolor[HTML]{9ecae1}37.29 \\
~+ OpenOrca & \cellcolor[HTML]{9ecae1}28.91 & \cellcolor[HTML]{9ecae1}41.91 & \cellcolor[HTML]{9ecae1}43.47 & \cellcolor[HTML]{9ecae1}49.87 & \cellcolor[HTML]{9ecae1}36.92 \\
\midrule
~+ Orignal Teacher & \cellcolor[HTML]{9ecae1}34.32 & \cellcolor[HTML]{9ecae1}43.06 & \cellcolor[HTML]{9ecae1}44.13 & \cellcolor[HTML]{9ecae1}52.13 & \cellcolor[HTML]{9ecae1}40.04 \\
~+ Rationale Only & \cellcolor[HTML]{9ecae1}37.43 & \cellcolor[HTML]{9ecae1}42.17 & \cellcolor[HTML]{6baed6}\textbf{47.47} & \cellcolor[HTML]{9ecae1}49.73 & \cellcolor[HTML]{9ecae1}41.30 \\
~+ Question Only & \cellcolor[HTML]{6baed6}\textbf{40.39} & \cellcolor[HTML]{9ecae1}43.85 & \cellcolor[HTML]{FDE9D9}41.73 & \cellcolor[HTML]{9ecae1}51.87 & \cellcolor[HTML]{9ecae1}42.78 \\
\midrule
~+ \ourmethod & \cellcolor[HTML]{6baed6}39.86 & \cellcolor[HTML]{6baed6}\textbf{44.45} & \cellcolor[HTML]{6baed6}46.13 & \cellcolor[HTML]{6baed6}\textbf{53.87} & \cellcolor[HTML]{6baed6}\textbf{43.44} \\
\bottomrule
\\
\end{tabular}
}
\caption{Accuracy (\%) of the student model Gemma-2B fine-tuned with different instruct-tuning datasets on Big-Bench-Hard under zero-shot and three-shot settings. Cells are colored blue if the method improves over Vanilla Gemma-2B, and orange if it declines.}
\label{tab:bbh-results}
\vspace*{-0.5cm}
\end{table}
\paragraph{Implementation}
In our experiments, we adopt Llama-3-70B-Instruction~\cite{metallama3} as the teacher model and Gemma-2B~\cite{banks2024gemma} as the student model.

In the \textit{Knowledge Elicitation} step,
1) \textit{Question Generation}:
We decode the question generation prompt in Table~\ref{tab:appendix-question-generation-prompt} with the aligned teacher at a temperature of $1.0$ until we achieve a total of $250$ draft questions per task.
2) \textit{Rationale Generation}:
For each draft question, the teacher model is prompted to generate $n=4$ rationales using the prompt in Table~\ref{tab:task-to-prompt} with diverse reasoning techniques.
Each draft question with its corresponding rationales forms a draft sample $(q, R_q)$.

In the \textit{Preference Collection} step, we collect the preference scores of questions and rationales through in-context learning on the preference set.
Specifically, we use the prompt template in Table~\ref{tab:pref-in-context} and decoding at a temperature $\tau=0$ to ensure the model's answer is deterministic.

In the \textit{Preference Alignment} step, we utilize preference scores to create the DPO dataset and align the teacher model with the student model's preferences.
For question generation, we select $50$ draft questions with the highest and lowest preference scores per task as preferred and dispreferred respectively, creating 50 DPO training examples at the question level. 
For rationale generation, the highest and lowest score rationales per question are selected similarly, resulting in $250$ DPO training examples at the rationale level per task. 
These datasets are combined to form the final DPO dataset. 
We employ Direct Preference Optimization (DPO) to align the models at a learning rate of $1e-7$, batch size of 16, for 1 epoch, incorporating a linear warm-up over 10\% of training steps followed by a cosine decay learning rate schedule.

Finally, we repeat the first step to generate tailored examples for BBH tasks and use them to Supervised Fine-Tune the student model with the same hyperparameters as baselines. All the experiments are conducted on 8 * NVIDIA A100 80G GPUs, which will cost about 1 day in total.

\subsection{Results}

Table~\ref{tab:bbh-results} shows the results of our method and baselines on the Big-Bench-Hard dataset under the zero-shot and three-shot in-context learning settings.
1) Compared to academic baselines: we can see that our method outperforms all the baselines including the baselines that simply imitate the reasoning process of the most powerful LLM (GPT-4-LLM), or mix multiple existing high-quality instruction-tuning datasets (Tulu-v2), and other two baselines that distill the teacher model with hand-crafted curations (OpenOrca and WizardLM).
2) Compared to the ablation baselines: our method outperforms both question-only and rationale-only baselines by a large margin, indicating that both the questions and rationales generated by the aligned teacher model are important for the student model to achieve better performance.
Besides, notice that the performance of the question-only baseline achieves better performance than the rationale-only baseline.
This interesting phenomenon indicates that in Knowledge Distillation for the language model, the quality of questions plays a more important role than the quality of rationales.

\section{Analysis on Generalization}
Although \ourmethod outperforms the baselines on the Big-Bench-Hard dataset, it is still unclear the generalization performance of \ourmethod.
Thus in this section, we present a thorough examination of its generalizability in the following two aspects:
1) Generalization of the fine-tuned student model.
2) Generalization of the aligned teacher model.
\subsection{Generalization of the Fine-tuned Student Model}
To investigate the generalization ability of the fine-tuned student model Gemma-2B, we conduct experiments on the following academic reasoning benchmarks:
1) PIQA~\cite{bisk2020piqa} is a physical commonsense reasoning dataset that is designed to test the model's ability to build, craft, or manipulate objects using everyday physical knowledge.
2) CSQA~\cite{talmor-etal-2019-commonsenseqa} is a question-answering benchmark targeting commonsense reasoning.
3) ARC-Easy and 4) ARC-Challenge~\cite{clark2018think} are multiple-choice question-answering benchmarks designed to test the model's ability to reason about scientific knowledge.
5) GSM8K~\cite{cobbe2021gsm8k} is a mathematical benchmark that is designed to test language models' ability in math and logic reasoning.


\begin{table}[t]
    \centering
    \label{tab:ood-results}
    \resizebox{0.9\textwidth}{!}{%
    \begin{tabular}{lccccc|c}
    \toprule
    \textbf{Models} & \textbf{PIQA} & \textbf{CSQA} & \textbf{ARC-E} & \textbf{ARC-C} & \textbf{GSM8K} & \textbf{Average} \\
    \midrule
    Vanilla Gemma-2B & 61.6 & 38.2 & 57.2 & 42.8 & 20.0 & 43.96 \\
    \midrule
    ~+ GPT-4-LLM & \cellcolor[HTML]{FDD7C1}60.2 & \cellcolor[HTML]{B4D3E6}42.8 & \cellcolor[HTML]{FDD7C1}55.8 & \cellcolor[HTML]{B4D3E6}46.2 & \cellcolor[HTML]{FDD7C1}18.6 & \cellcolor[HTML]{D8E8F2}44.72 \\
    ~+ Tulu-v2 & \cellcolor[HTML]{92BFDB}\textbf{65.2} & \cellcolor[HTML]{B4D3E6}42.8 & \cellcolor[HTML]{9EC7DF}62.2 & \cellcolor[HTML]{6FAAD4}49.6 & \cellcolor[HTML]{FDD7C1}18.8 & \cellcolor[HTML]{9EC7DF}47.72 \\
    ~+ WizardLM & \cellcolor[HTML]{C5DDEC}63.6 & \cellcolor[HTML]{C5DDEC}41.4 & \cellcolor[HTML]{FEF1EB}56.4 & \cellcolor[HTML]{6FAAD4}49.2 & \cellcolor[HTML]{9EC7DF}26.2 & \cellcolor[HTML]{9EC7DF}47.36 \\
    ~+ OpenOrca & \cellcolor[HTML]{C5DDEC}64.6 & \cellcolor[HTML]{9EC7DF}\textbf{43.6} & \cellcolor[HTML]{C5DDEC}61.0 & \cellcolor[HTML]{9EC7DF}48.4 & \cellcolor[HTML]{C5DDEC}23.6 & \cellcolor[HTML]{9EC7DF}48.24 \\
    \midrule
    ~+ \ourmethod & \cellcolor[HTML]{C5DDEC}63.4 & \cellcolor[HTML]{9EC7DF}42.9 & \cellcolor[HTML]{9EC7DF}\textbf{63.6} & \cellcolor[HTML]{348AC7}\textbf{57.2} & \cellcolor[HTML]{348AC7}\textbf{32.0} & \cellcolor[HTML]{348AC7}\textbf{51.82} \\
    \bottomrule
    \\
    \end{tabular}
    }
    \caption{Accuracy (\%) of the student model Gemma-2B fine-tuned with \ourmethod and academic baselines on out-of-domain reasoning benchmarks. 
    Cells are colored blue if the method improves over Vanilla Gemma-2B, and orange if it declines.}
    \label{tab:ood-results}
    \vspace*{-0.5cm}
\end{table}

Table~\ref{tab:ood-results} shows the results of our method and baselines on the five benchmarks.
On average, our method outperforms the best baseline by $1.5\%$ in accuracy.
The results show that our method helps the student model Gemma-2B to achieve better generalization ability on reasoning tasks.
In some tasks like PIQA and CSQA, the performance of our method is not the best but still competitive.
While in more challenging tasks like ARC-Challenge and GSM8K, which are both sampled from real-world grade school exams, our method outperforms all the baselines.
This phenomenon reveals for more challenging tasks, more carefully curated examples are needed in knowledge distillation.
Compared to the heuristic baselines, the tailored examples generated by the aligned teacher model are more effective in these tasks as they can better capture the student model's preferences.

\subsection{Generalization of the Aligned Teacher Model}
Due to the high computational cost of preference collection and alignment, it is desirable that the aligned teacher model can generate tailored examples for unseen tasks and unseen student models.

\noindent{\bf Generalization Across Tasks}\quad  To investigate whether our aligned teacher model is capable of generating tailored examples for student models on unseen tasks, we conduct experiments on the PIQA, ARC-EASY, ARC-Challenge, and GSM8K benchmarks.
\begin{wraptable}{r}{0.55\textwidth}
    \centering
    \scalebox{0.85}{
        \begin{tabular}{@{}lcccc|c@{}}
        \toprule
        \textbf{Models}                & \textbf{PIQA}  & \textbf{ARC-E} & \textbf{ARC-C} & \textbf{GSM8K} & \textbf{Average} \\
        \midrule
        \multicolumn{6}{c}{\textit{zero-shot results}}                                      \\
        \midrule
        Original  & 62.4 & 62.4    & 51.0          & 22.2  & 49.5   \\
        Aligned   & 65.6 & 59.4    & 51.2          & 25.6  & 50.45  \\
        Delta     & \textcolor{teal}{$\uparrow$ 3.2} & \textcolor{red}{$\downarrow$ 3.0}    & \textcolor{teal}{$\uparrow$ 0.2}          & \textcolor{teal}{$\uparrow$ 3.4}  & \textcolor{teal}{$\uparrow$ 0.95}  \\
        \midrule
        \multicolumn{6}{c}{\textit{three-shot results}}                                      \\
        \midrule
        Original  & 65.0 & 64.2    & 51.8          & 23.4  & 51.1   \\
        Aligned   & 66.6 & 65.2    & 52.6          & 24.8  & 52.3   \\
        Delta     & \textcolor{teal}{$\uparrow$ 1.6} & \textcolor{teal}{$\uparrow$ 1.0}    & \textcolor{teal}{$\uparrow$ 0.8}          & \textcolor{teal}{$\uparrow$ 1.4}  & \textcolor{teal}{$\uparrow$ 1.2}   \\
        \bottomrule
        \end{tabular}
    }
    \caption{Accuracy (\%) of the student model Gemma-2B fine-tuned with training examples generated by the original teacher model and the aligned teacher model on out-of-domain reasoning benchmarks.}
    \label{tab:task-generalization}
    \vspace*{-0.5cm}
\end{wraptable}

We repeat Step 1 \textit{Knowledge Elicitation} to generate training examples on the four benchmarks.
Specifically, we guide both original and aligned teacher models to generate $2500$ training examples for each unseen benchmark. 
We train the student model Gemma-2B with the generated examples 
on the four unseen benchmarks separately
with the same hyperparameters mentioned in Section~\ref{sec:experiments}.

Results in Table~\ref{tab:task-generalization} show that the student model fine-tuned with training examples generated by the aligned teacher model outperforms the student model fine-tuned with training examples generated by the original teacher model.
This finding suggests that through preference alignment, the aligned teacher model gains a deeper understanding of the student model’s preferences, which helps with unseen tasks. 
This enhanced understanding of the teacher model makes it easier to distill the specific abilities of the aligned teacher model to the student model compared to the original teacher model.

\begin{wraptable}{r}{0.55\textwidth}
    \centering
    \scalebox{0.85}{
        \begin{tabular}{@{}lccc@{}}
        \toprule
        \textbf{Models}                & \textbf{Gemma-7B}  & \textbf{Qwen1.5-1.8B} & \textbf{CodeGemma-2B} \\
        \midrule
        \multicolumn{4}{c}{\textit{zero-shot results}}                                      \\
        \midrule
        Original              & 49.0      & 36.3       & 38.7         \\
        Aligned               & 49.0      & 39.4       & 41.2         \\
        Delta                 & 0.0   & \textcolor{teal}{\(\uparrow\) 3.1} & \textcolor{teal}{\(\uparrow\) 2.5} \\
        \midrule
        \multicolumn{4}{c}{\textit{three-shot results}}                                      \\
        \midrule
        Original              & 51.2      & 38.2       & 39.8         \\
        Aligned               & 51.0      & 40.3       & 42.7         \\
        Delta                 & \textcolor{red}{\(\downarrow\) 0.2} & \textcolor{teal}{\(\uparrow\) 2.1} & \textcolor{teal}{\(\uparrow\) 2.9} \\
        \bottomrule
        \end{tabular}
    }
    \caption{Accuracy (\%) of the different student models fine-tuned with training examples generated by the original teacher model and the aligned teacher model on Big-Bench-Hard.}
    \label{tab:student-generalization}
    \vspace*{-0.5cm}
\end{wraptable}

\noindent{\bf Generalization Across Student Models}\quad To investigate the generalization ability of the aligned teacher model across unseen student models, we conduct experiments on the Big-Bench-Hard dataset.
Specifically, we adopt Gemma-7B as the student model from the same family with a different capacity, 
Qwen1.5-1.8B and CodeGemma-2B as the student models with the same capacity but from different families or different domains.
We train the student models with the training examples generated by the original teacher model and the aligned teacher model under the same hyperparameters mentioned in Section~\ref{sec:experiments}.

Table~\ref{tab:student-generalization} shows the results of the student models trained with the original teacher model and the aligned teacher model, denoted as \textit{Original} and \textit{Aligned} respectively.
The results show that in Qwen1.5-1.8B and CodeGemma-2B, the models from different families or different domains share similar parameter capacity with the Gemma-2B, and achieve better performance with the tailored examples generated by the aligned teacher model than the original teacher model.
In Gemma-7B, the model from the same family with a larger parameter capacity does not present a promising improvement.
This indicates that the language model with a similar parameter capacity shares similar preferences in training examples.
This finding suggests that the aligned teacher model can be applied to different student models as long as they share similar parameter capacities.

%









\section{Related Work}
\label{sec:related-work}

\paragraph{Knowledge Distillation for LLMs}
Knowledge distillation is a widely used technique to transfer knowledge from a large teacher model to a smaller student model~\cite{hinton2015distilling}.
Recently, the most powerful LLMs are computationally expensive highlighting the need for knowledge distillation to transfer advanced capabilities from large LLMs to lightweight models~\cite{xu2023wizardlm,mitra2023orca2,timiryasov2023baby}.
There are two main steps in knowledge distillation~\cite{xu2024survey}: (1) Knowledge Elicitation, which generates training examples for the student model, and (2) Distillation Algorithms, which centers on injecting the elicited knowledge, namely training examples, into the student model.
Although there are some other distillation algorithms~\cite{wan2024knowledge}, 
Supervised Fine-Tuning~\cite{brown2020language} has been the de facto standard for knowledge distillation due to its simplicity and effectiveness.
Most work focus on knowledge elicitation, specifically on how to curate training data from LLMs~\cite{xu2024survey}
\citet{hsieh-etal-2023-step-by-step,mukherjee2023orca} proposed to guide the teacher language model to generate not only the final answer but also the reasoning process with chain-of-thought~\cite{wei2022chain,yao2022react}, which provides more information for the student model to imitate.
Since it is easy to synthesize large-scale text from the teacher model, the key challenge lies in the quality and diversity control of the generated examples.
\citet{wang2022self} utilize the ROUGE-L similarity~\cite{lin-2004-rouge} to filter out the redundant examples to ensure the diversity.
Inspired by the high quality of the textbook, \citet{li2023phi1.5} proposed to synthesize ``textbook-like'' text from the teacher model.
\citet{mitra2023orca2} contends that different tasks should be combined with different reasoning strategies when generating rationales of examples.
All these works have advanced in curating training examples, but they overlook aligning the teacher model with responses from the student model to refine the training data, referred to as ``responsive teaching'' in pedagogy~\cite{gay2000culturally}, which effectively lowers the learning barriers for students~\cite{hattie2007power}.

\paragraph{Alignment of LLMs}
Large language models (LLMs) trained on large-scale corpora have shown remarkable performance on various natural language processing tasks~\cite{achiam2023gpt4,touvron2023llama2,chowdhery2023palm,jiang2023mistral}.
Despite the success, these models may not be aligned with the needs of human users generating text with biases, factual inaccuracies, or inappropriate content~\cite{wang2023aligning}.
Many works have been proposed to align LLMs with human preferences.
Supervised Fine-Tuning~\cite{brown2020language} is the basic approach to align LLMs with human preferences, which aligns the model by minimizing the cross-entropy loss between the model's prediction and the human-labeled data.
Reinforcement learning from human preference (RLHF)~\cite{christiano2017deep} prosed to optimize for maximum reward operates by engaging with a reward model trained using the Bradley-Terry model~\cite{bradley1952rank,bong2022generalized} through reinforcement algorithm Proximal Policy Optimization (PPO)~\cite{schulman2017proximal}.
While RLHF exhibits remarkable performance compared to supervised fine-tuning, it faces challenges such as reward hacking~\cite{liu2023statistical}.
Direct Preference Optimization (DPO)~\cite{rafailov2023dpo} has emerged as a promising alternative to RLHF, which does not require a reward model as a proxy.
It recasts the alignment formulation as a simple loss function that can be optimized directly on a dataset of preference pairs ${(x,y_w,y_l)}$, where $x$ is prompt and $y_w$ and $y_l$ are the preferred and dispreferred responses, respectively.
Since DPO is RL-free and does not require a reward model, its training process is more stable than RLHF.~\cite{wang2023aligning}
In this work, we leverage the DPO to align the teacher model with the student model.
\section{Conclusion}
In this study, we propose \ourmethod, a novel framework in Knowledge Distillation that aligns the teacher language model with the student language model's preferences to generate tailored training examples for Knowledge Distillation, which is inspired by responsive teaching in pedagogy, which effectively lowers the learning barriers for students.
Our framework consists of three main steps: \textit{Knowledge Elicitation, Preference Collection, and Preference Alignment}.
Extensive experiments on various academic benchmarks show that our method outperforms existing instruction-tuning datasets distilled from powerful LLMs by a large margin.
Moreover, we investigate the generalization of the aligned teacher model, showing that the aligned teacher model can be generalized to other reasoning benchmarks and different student models.







\bibliography{custom}


\appendix
\newpage
\section*{Appendix}

\section{Limitations}
\label{sec:limitations}
Although most of the data used in \ourmethod are automatically generated by the teacher model, there still needs some manual efforts to construct the prompts and collect the preference scores.
Specifically, there are two main limitations in the current implementation of \ourmethod:
First, in draft rationale generation, to elicit diverse and high-quality rationales, a set of carefully designed prompts is required. 
In this work, we hand-crafted the system prompts for each task with different reasoning techniques by the paper authors themselves.
Totally, we constructed 4 prompts for each task as presented in Table~\ref{tab:task-to-prompt}.
Recently, \citet{wang2024chain} proposed to Chain-of-Thought Decoding (CoT-Decoding) to uncover reasoning processes of questions from language models without prompts.
In the future, we will explore the possibility of using CoT-Decoding to generate a reasoning process automatically.
Second, in preference collection, a set of labeled examples consisting of questions and answers is required to act as the validation set and preference set.
Preference scores are collected on these labeled question-answer pairs to measure the preference of the student model towards the draft questions and rationales.
In this work, we simply reuse the data from the original Big-Bench dataset as the validation set.
In the future, we will explore the possibility of directly measuring the preference through the internal states of the student model~\cite{kadavath2022language}.

\section{Broader Impacts}
\label{sec:broader-impacts}
The possible broader impacts of this work lie in the hallucination of language models.
Language models have been shown to generate biased and harmful content~\cite{bender2021dangers, zhang2023siren}.
In this work, we focus on the enhancement of the student model's specific capability, such as reasoning, which is not directly relevant to the generation of harmful content.
It is worth noting that the user should be cautious when using the student model or the teacher model to generate text, especially when the generated text is used in critical applications such as medical diagnosis or legal advice.

\section{Insights for Generating Rationale for Small Language Models}
\label{sec:insights}
\begin{figure}[h]
    \centering
    \begin{minipage}[t]{0.50\textwidth}
        \centering
        \includegraphics[width=\textwidth]{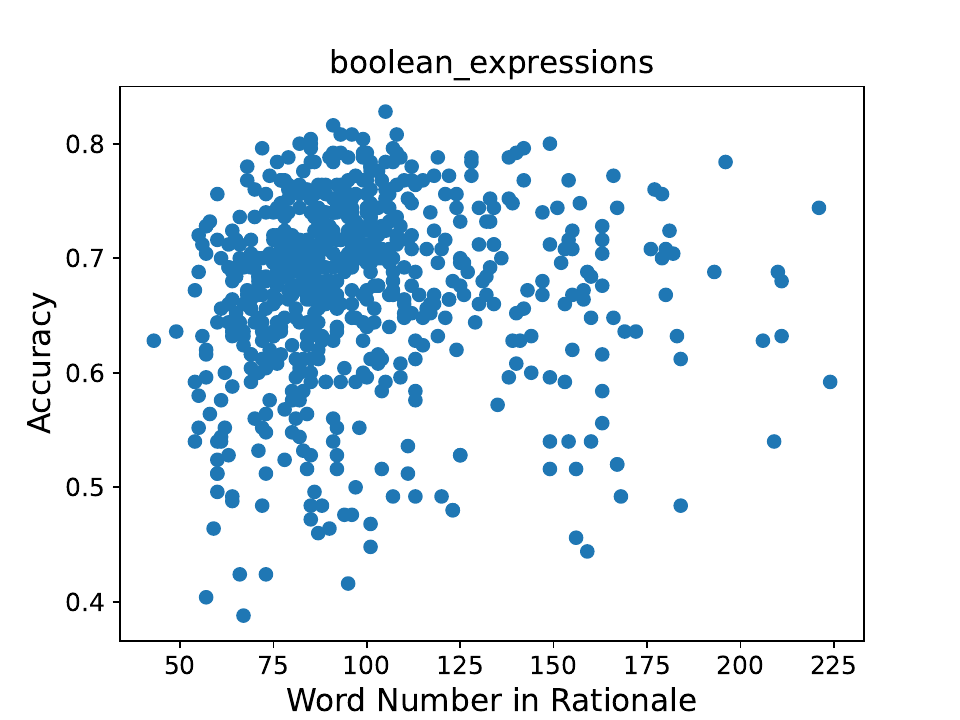}
    \end{minipage}%
    \hfill
    \begin{minipage}[t]{0.50\textwidth}
        \centering
        \includegraphics[width=\textwidth]{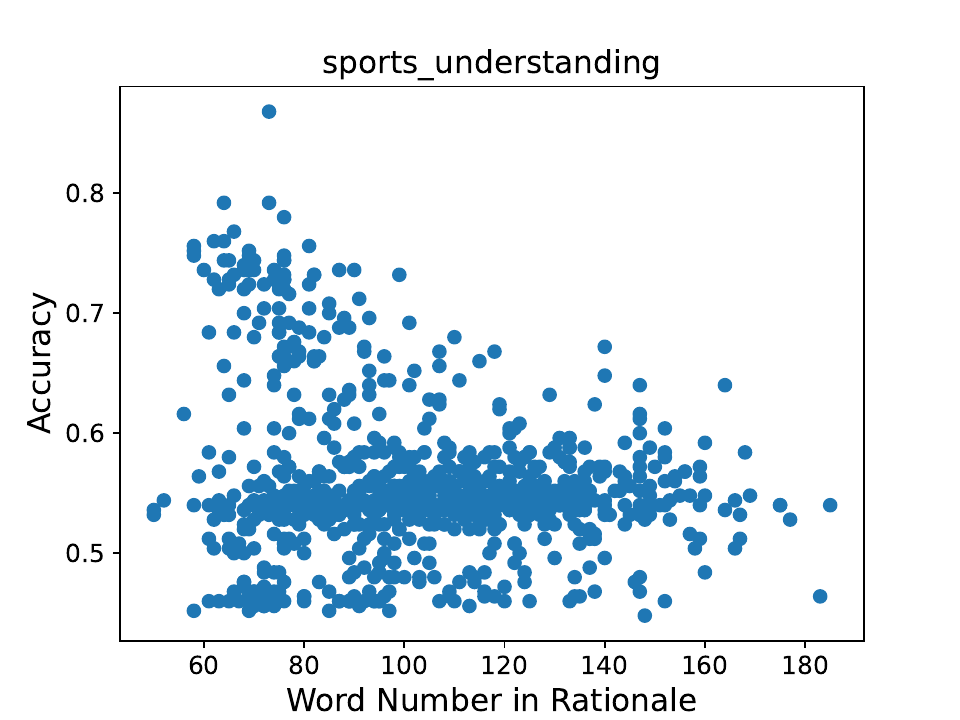}
    \end{minipage}
    \caption{The relationship between the word number of the rationale and the one-shot in-context learning accuracy of the Gemma-2B on boolean expressions and sports understanding tasks.}
    \label{fig:insight-pic1}
\end{figure}

In Step 2 \textit{Preference Collection}, we collect the preference of small language models, such as Gemma-2B, towards the draft examples through one-shot in-context learning on the preference set.
Through careful analysis of the collected preference scores, we conclude with two insights for generating a tailored example for language models with limited capacity:

\paragraph{Insight 1: The more detailed the rationale does not necessarily mean the better the performance of the small language model.}
In previous works~\cite{hsieh-etal-2023-step-by-step,mukherjee2023orca}, the teacher model is guided to generate a detailed reasoning process with chain-of-thought~\cite{wei2022chain,yao2022react} to provide more information for the small language model to imitate.
However, after taking a closer look at the preference scores, 
we find that there is no significant linear correlation between the length of the rationale and the preference score, namely the accuracy of the small language model within one-shot in-context learning.
Figure~\ref{fig:insight-pic1} shows the relationship between the word number of the rationale and the one-shot in-context learning accuracy of the small language model on boolean expressions and sports understanding tasks.

As shown in Figure~\ref{fig:insight-pic1}, the rationale with most words does not necessarily mean the best performance of the small language model, while the rationale with moderate words (75-100 words in Boolean Expressions and 50-75 words in Sports Understanding) achieves better results.
This indicates that \textit{a rationale with complete and concise reasoning steps is more beneficial for the small language model to learn}, compared to a long meticulous rationale.

The reasons behind this are twofold: 
First, the \textit{Lost-in-the-Middle Phenomenon}~\cite{liu2024lost} of language models.
Language models may lose information when the input context is too long.
For limited-capacity language models, when example rationales are too long, the small language model may get lost in the narrative and forget the question to solve.
Second, the \textit{Repetition Problem}~\cite{welleck2019neural} of language models.
In long meticulous rationales, the teacher model may repeat the same step multiple times, for example acting as a stack machine when solving a math problem.
For limited-capacity language models, this repeated step may cause the small language model to get stuck in a loop and repeat the same step infinitely.

\paragraph{Insight 2: Although different questions in the same task prefer different reasoning strategies, the small language model prefers consistent reasoning strategies for one task in Supervised Fine-tuning.}
\begin{wraptable}{r}{0.40\textwidth}
    \centering
    \scalebox{1}{
    \begin{tabular}{@{}lcc@{}}
    \toprule
    \textbf{Method} & \textbf{BBH} \\
    \midrule 
    Vanilla Gemma-2B & 35.57 \\
    \midrule
    ~+ Randomly Selected & 39.94 \\ 
    ~+ Most Preferred & 40.04  \\
    ~+ Task Consistent & 42.12 \\
    ~+ Aligned Teacher & 42.78 \\
    \bottomrule
    \end{tabular}
    }
    \caption{Accuracy (\%) of the Gemma-2B fine-tuned with different training examples on Big-Bench-Hard.}
    \label{tab:insight-2}
\end{wraptable}

The diversity of tests in the training corpus is crucial for the pretraining stage of language models~\cite{liu2019roberta}.
To investigate the impact of the diversity of reasoning strategies in the training dataset, we fine-tuned the Gemma-2B with four different training datasets on Big-Bench-Hard.
1) \textit{Randomly Selected}: the rationale of each question is randomly selected from different reasoning strategies.
2) \textit{Most Preferred}: the rationale of each question is selected based on the highest preference score.
3) \textit{Task Consistent}: all rationales are selected from within the same reasoning strategy for one task.
4) \textit{Aligned Teacher}: the rationales are generated by the aligned teacher model with the small language model's preference.
All the questions in these four training datasets are generated by the original teacher model to ensure fairness.
Table~\ref{tab:insight-2} presents the performance of the small language model fine-tuned with four different training datasets on Big-Bench-Hard.
The results present an interesting phenomenon: the performance of Gemma-2B fine-tuned with the \textit{Most Preferred} dataset is similar to the \textit{Randomly Selected} dataset, while \textit{Task Consistent} and \textit{Aligned Teacher} datasets both outperform the other two datasets by a decent margin.
This indicates that \textit{in the fine-tuning stage, especially when we try to enhance one or two specific capabilities of the small language model, for example, reasoning, the consistent reasoning strategies are more beneficial.}
The reason behind this is that the small language model may get confused when the reasoning strategies are too diverse for one task, while the consistent reasoning strategies provide clear guidance for the small language model to imitate for specific capability enhancement.

\section{Big-Bench-Hard Category}
\label{sec:bbh-category}
We split the Big-Bench-Hard tasks into four categories based on the capabilities required by the tasks: (1) \textit{logical reasoning}, (2) \textit{commonsense reasoning}, (3) \textit{math reasoning} and (4) \textit{world knowledge}, denoted as \textit{BBH-Logical}, \textit{BBH-Commonsense}, \textit{BBH-Math} and \textit{BBH-Knowledge} respectively.
Table~\ref{tab:0-bbh-category} shows the detailed information of the tasks in Big-Bench-Hard.
\newpage
\begin{longtable}{lp{9cm}c}
\toprule
\textbf{Category} & \textbf{Task} & \textbf{Quantity} \\
\midrule
\multirow{4}{*}{BBH-Logic} & dyck\_languages, formal\_fallacies\_syllogisms\_negation, logical\_deduction, navigate, object\_counting, penguins\_in\_a\_table, temporal\_sequences, tracking\_shuffled\_objects, web\_of\_lies, word\_sorting & \multirow{4}{*}{10} \\
\midrule
\multirow{2}{*}{BBH-CS} & causal\_judgment, date\_understanding, disambiguation\_qa, hyperbaton, reasoning\_about\_colored\_objects, ruin\_names, snarks & \multirow{2}{*}{7} \\
\midrule
\multirow{1}{*}{BBH-Math} & boolean\_expressions, geometric\_shapes, multistep\_arithmetic & \multirow{1}{*}{3} \\
\midrule
\multirow{2}{*}{BBH-Knowl} & movie\_recommendation, salient\_translation\_error\_detection, sports\_understanding & \multirow{2}{*}{3} \\
\bottomrule
\\
\caption{Big-Bench-Hard task categories.}
\label{tab:0-bbh-category}
\end{longtable}

\section{2P-IRT Model and its Consistency}
\label{sec:consistency}

To find out the top 40 most discriminative questions to form the preference set, we first use the 2P-IRT model to estimate parameters. The 2P-IRT model involves two main parameters: difficulty parameter and discrimination parameter. The difficulty parameter measures the difficulty level of a question, representing the probability of an LLM reaching or exceeding that level in order to answer the question correctly. The discrimination parameter measures the question's ability to differentiate between different levels of ability, higher discrimination parameters indicate greater item discrimination. Then we use the MLE method to iteratively optimize the parameter estimates. The likelihood function is maximized by adjusting the parameter values step by step. Finally, we choose the top 40 questions with the highest discrimination parameters. 

To evaluate the consistency between the preference score collected from the preference set and the original validation set, we conducted an experiment where we selected one task from each category and generated 10 draft samples \((q, R_q)\) for each task. For each draft sample, we computed the preference scores of each rationale in \(R_q\) against both sets. The top-rated rationale aligned in 93.3\% of cases across both sets, and the lowest-rated had a consistency of 86.7\%. These results affirm the reliability of using the preference set to collect scores.

\section{Repeating \textit{Knowledge Elicitation} to Generate Training Examples}
\label{sec:repeating}
After the \textit{Preference Alignment} step, we need to repeat the \textit{Knowledge Elicitation} step to generate training examples for the student model with the aligned teacher model.
Specifically, first, we prompt the aligned teacher model with question generation prompts as shown in Table~\ref{tab:appendix-question-generation-prompt} to generate draft examples.
We will decode the question generation prompts with decoding temperature $1.0$ multiple times until we get enough draft questions.
Then, for each question, we prompt the aligned teacher model with naive \textit{Step-by-Step} prompts as shown in Table~\ref{tab:rationale-prompt} 
to generate rationales with decoding temperature $0.0$.
\begin{table}[h]
    \centering

    \begin{tabular}{p{0.85\linewidth}}
        \toprule
        Task Description: \{Task Description\} \\
        Question: \{Question\} \\
        Answer: Let's think step by step. \{to be completed by the teacher model\} \\
        \bottomrule
        \\
    \end{tabular}    
    \caption{
        Naive \textit{Step-by-Step} prompt for the teacher model.
    }
    \label{tab:rationale-prompt}
\end{table}

\section{Direct Preference Optimization Details}
\label{sec:dpo}
In the \textit{Preference Alignment} step, we align the teacher model with the student model's preferences through Direct Preference Optimization (DPO)~\cite{rafailov2023dpo}.
Here we provide the details of constructing the DPO dataset

\paragraph{DPO dataset for Question Generation}
For each task, we construct the DPO dataset for question generation using the following template,
\begin{table}[ht]
    \centering
    \begin{tabular}{l|l}
        \toprule
        Input Prompt $x$: &  \{ The Question Generation Prompt \} \\
        \midrule
        Preferred Response $y_w$: & \{Preferred Question\} \\
        Dispreferred Response $y_l$: & \{Dispreferred Question\} \\
        \bottomrule
        \\
    \end{tabular}    
    \caption{
        DPO dataset template for question generation.
    }
    \label{tab:dpo-question}
\end{table}
The question generation prompt is constructed based on the task description and seed questions, detailed in Table~\ref{tab:appendix-question-generation-prompt}.
The preferred response is randomly selected from the draft questions with top-25\% preference scores, while the dispreferred response is randomly selected from the draft questions with bottom-25\% preference scores.

\paragraph{DPO dataset for Rationale Generation}
For each draft question, we construct the DPO dataset for rationale generation using the following template,
\begin{table}[ht]
    \centering
    \begin{tabular}{l|l}
        \toprule
        Input Prompt $x$: &  \{ The Naive \textit{Step-by-Step} Prompt \} \\
        \midrule
        Preferred Response $y_w$: & \{Preferred Rationale\} \\
        Dispreferred Response $y_l$: & \{Dispreferred Rationale\} \\
        \bottomrule
        \\
    \end{tabular}    
    \caption{
        DPO dataset template for rationale generation.
    }
    \label{tab:dpo-rationale}
\end{table}
The naive \textit{Step-by-Step} prompt is constructed based on the task description and the draft question, detailed in Table~\ref{tab:rationale-prompt}.
The idea of using naive \textit{Step-by-Step} prompts is insipred by \textit{Prompt Erasure}~\cite{mitra2023orca2}, which lets the teacher model learn to select the most suitable reasoning strategy to generate tailored training examples for the student model.
The preferred response is the draft rationale to the draft question with the highest preference scores, while the dispreferred response is the draft rationale with the lowest preference scores.

Finally, we mix the DPO dataset for question generation and rationale generation together to train the teacher model with DPO.

\section{Question Generation Prompt}
We provide some prompts for guiding the teacher model to generate questions in the \textbf{Knowledge Elicitation} step.

{
\small
\begin{longtable}{cp{10cm}}
\toprule
\textbf{Task} & \textbf{Prompt} \\
\midrule
boolean\_expressions & boolean expressions is a task whose description is as follows: Evaluate the result of a random Boolean expression. \\ 
& I will provide you 3 example questions about this task in the following. Based on the following example, I want you to generate 5 more questions about boolean expressions with brainstorming. \\ 
& You may not stick to context, but the question should be related to boolean expressions. \\ 
& The output should be like this:\\
& <Question\_x><input>...</input><target>...</target></Question\_x> \\ 
& x can be any number, but it should be unique. \\ 
& \\
& <Question\_1><input>not ( ( not not True ) ) is</input> <target>False</target></Question\_1> \\ 
& <Question\_2><input>True and False and not True and True is</input> <target>False</target></Question\_2> \\ 
& <Question\_3><input>not not ( not ( False ) ) is</input> <target>True</target></Question\_3> \\
\midrule
causal\_judgment & causal judgment is a task whose description is as follows: Answer questions about causal attribution. \\ 
& I will provide you 3 example questions about this task in the following. Based on the following example, I want you to generate 5 more questions about causal judgment with brainstorming. \\ 
& You may not stick to context, but the question should be related to causal judgment. \\ 
& The output should be like this:\\
& <Question\_x><input>...</input><target>...</target></Question\_x> \\ 
& x can be any number, but it should be unique. \\ 
& \\
& <Question\_1><input>How would a typical person answer each of the following questions about causation? \\ 
& Frank T., had an ongoing dispute with his neighbor over a stretch of land and one day decided to shoot his neighbor in the body. Frank T. had no experience with guns, his hand slipped on the barrel of the gun, and the shot went wild. Nonetheless, the bullet bounced off a large boulder several feet away and hit the neighbor's body, causing significant injury. Did Frank T. intentionally shoot his neighbor in the body? \\ 
& Options: \\ 
& - Yes \\ 
& - No</input><target>No</target></Question\_1> \\ 
& <Question\_2><input>How would a typical person answer each of the following questions about causation? \\ 
& Suzy and Billy are working on a project that is very important for our nation's security. The boss tells them both: "Be sure that you are here at exactly 9 am. It is absolutely essential that you arrive at that time." Both Billy and Suzy arrive at 9 am. As it happens, there was a motion detector installed in the room where they arrived. The motion detector was set up to be triggered if at least one person appeared in the room at the same time. So the motion detector went off. Did Billy cause the motion detector to go off? \\ 
& Options: \\ 
& - Yes \\ 
& - No</input><target>Yes</target></Question\_2> \\ 
& <Question\_3><input>How would a typical person answer each of the following questions about causation? \\ 
& George and his sister Lena reunite at their parent's house for Thanksgiving. Whereas George just got into medical school, Lena is unhappy in her marriage and recently lost her job. Over the course of the day, George and Lena get into a number of heated arguments. Later in the afternoon, they play a game of darts. They split the first two games, and the third game was close until the end. Who will win comes down to George's last shot. If he hits a high point region, he wins; if he hits a low point region, Lena wins. George thinks of the difficult time Lena is having, and he really wants to let her win. He aims the dart at the low point region. He sets up his shot and the dart lands in the low point region. After his shot, Lena wins the game and is very happy. Did George hit the low point region intentionally? \\ 
& Options: \\ 
& - Yes \\ 
& - No</input><target>Yes</target></Question\_3> \\
\midrule
date\_understanding & date understanding is a task whose description is as follows: Infer the date from context. \\ 
& I will provide you 3 example questions about this task in the following. Based on the following example, I want you to generate 5 more questions about date understanding with brainstorming. \\ 
& You may not stick to context, but the question should be related to date understanding. \\ 
& The output should be like this:\\
& <Question\_x><input>...</input><target>...</target></Question\_x> \\ 
& x can be any number, but it should be unique. \\ 
& \\
& <Question\_1><input>Today is Christmas Eve of 1937. What is the date 10 days ago in MM/DD/YYYY? \\ 
& Options: \\ 
& (A) 12/14/2026 \\ 
& (B) 12/14/1950 \\ 
& (C) 12/14/2007 \\ 
& (D) 12/14/1937 \\ 
& (E) 07/14/1938 \\ 
& (F) 12/14/1988</input><target>(D)</target></Question\_1> \\ 
& <Question\_2><input>Tomorrow is 11/12/2019. What is the date one year ago from today in MM/DD/YYYY? \\ 
& Options: \\ 
& (A) 09/04/2018 \\ 
& (B) 11/11/2018 \\ 
& (C) 08/25/2018 \\ 
& (D) 11/02/2018 \\ 
& (E) 11/04/2018</input><target>(B)</target></Question\_2> \\ 
& <Question\_3><input>Jane and John married on Jan 2, 1958. It is their 5-year anniversary today. What is the date tomorrow in MM/DD/YYYY? \\ 
& Options: \\ 
& (A) 01/11/1961 \\ 
& (B) 01/03/1963 \\ 
& (C) 01/18/1961 \\ 
& (D) 10/14/1960 \\ 
& (E) 01/03/1982 \\ 
& (F) 12/03/1960</input><target>(B)</target></Question\_3> \\
\midrule
movie\_recommendation & movie recommendation is a task whose description is as follows: Recommend movies similar to the given list of movies. \\ 
& I will provide you 3 example questions about this task in the following. Based on the following example, I want you to generate 5 more questions about movie recommendations by brainstorming. \\ 
& You may not stick to context, but the question should be related to movie recommendations. \\ 
& The output should be like this:\\
& <Question\_x><input>...</input><target>...</target></Question\_x> \\ 
& x can be any number, but it should be unique. \\ 
& \\
& <Question\_1><input>Find a movie similar to Star Wars Episode IV - A New Hope, Indiana Jones and the Last Crusade, Star Wars Episode V - The Empire Strikes Back, The Big Lebowski: \\ 
& Options: \\ 
& (A) Tetsuo \\ 
& (B) the Ironman \\ 
& (C) The Princess Bride \\ 
& (D) The Barkley Marathons The Race That Eats Its Young \\ 
& (E) Bug</input><target>(C)</target></Question\_1> \\ 
& <Question\_2><input>Find a movie similar to Twister, The Silence of the Lambs, Independence Day, Braveheart: \\ 
& Options: \\ 
& (A) They Shoot Horses \\ 
& (B) Don't They \\ 
& (C) Forrest Gump \\ 
& (D) The Salton Sea \\ 
& (E) Extreme Days</input><target>(C)</target></Question\_2> \\ 
& <Question\_3><input>Find a movie similar to Minority Report, Total Recall, Inside Out, Forrest Gump: \\ 
& Options: \\ 
& (A) Phenomena \\ 
& (B) Lilting \\ 
& (C) Catwoman \\ 
& (D) Edge of Tomorrow</input><target>(D)</target></Question\_3> \\
\bottomrule
\\
\caption{Prompts for question generation.}
\label{tab:appendix-question-generation-prompt}
\end{longtable}
}

\section{Rationale Generation Prompt for Different Reasoning Techniques}
We provide some prompts for guiding the teacher model to generate rationale with different reasoning techniques in the \textbf{Knowledge Elicitation} step.
{
\small
\begin{longtable}{cp{9cm}}
\toprule
\textbf{Task} & \textbf{Prompt} \\
\midrule
\multirow{4}{*}{boolean\_expressions} & Image you are an expert in Boolean expression evaluation. Now you will be given a random Boolean expression, you should first evaluate the expressions inside brackets, then follow the order of operations from highest priority to lowest priority namely "not", "and", "or", respectively, and finally evaluate the result of the random Boolean expression. Remember you should output your final answer in the end like <ans>True</ans> or <ans>False</ans> \\
\cmidrule{2-2}
~ & You are an expert in Math. Given a random Boolean expression, you should first recall the rules of Boolean algebra and then evaluate the expression step by step. Finally, you should provide the result of the expression. Remember you should output your final answer in the end like <ans>True</ans> or <ans>False</ans> \\
\cmidrule{2-2}
~ & Evaluate the result of a random Boolean expression. Remember you should output your final answer in the end like <ans>True</ans> or <ans>False</ans> \\
\cmidrule{2-2}
~ & Given you are a binary classification question, solve the question step by step as follows: 1. Read the question and options 2. Find the best option among the remaining ones. Remember you should output your final answer in the end like <ans>True</ans> or <ans>False</ans> \\
\midrule
\multirow{4}{*}{causal\_judgment} & I want you to act as a judge in a causal judgment process. There are 3 key points you should consider when assessing whether an action was intentional or not. 1. Intentionality and Outcome: When evaluating an action, consider the actor's intentions and the sequence of events leading to the outcome. If the actor clearly intended the outcome and acted to bring it about, the action is intentional. 2. Accidents and Unintended Consequences: If the outcome resulted from an accident or slip, and not from a deliberate act aimed at causing that specific outcome, it is considered unintentional. Unexpected or uncontrollable factors can influence the final result, diverging from the actor's original intent. 3. Contribution to Outcome: In cases where multiple factors contribute to an outcome, analyze each factor's role in producing the final effect. If an agent's actions directly contribute to triggering an event, they can be considered a cause, even if other factors are also involved. Carefully assess the situation and the actor's mindset to make a fair and accurate judgment. Based the three key points, given one scenario, you should first analyze the situation and the actor's mindset. Then, decide which key point is the most relevant to the scenario. Finally, you should conclude by answering the question of whether an action was intentional or not. Remember you should output your final answer in the end like <ans>True</ans> or <ans>False</ans> \\
\cmidrule{2-2}
~ & You are given a scenario and asked to figure out whether the person in the scenario intentionally caused the outcome. Use your common sense and the information given in the scenario to answer the question step by step. Remember you should output your final answer in the end like <ans>True</ans> or <ans>False</ans> \\
\cmidrule{2-2}
~ & Answer questions about causal attribution. Remember you should output your final answer in the end like <ans>True</ans> or <ans>False</ans> \\
\cmidrule{2-2}
~ & Given you are a binary classification question, solve the question step by step as follows: 1. Read the question and options 2. Find the best option among the remaining ones. Remember you should output your final answer in the end like <ans>True</ans> or <ans>False</ans>\\
\midrule
\multirow{4}{*}{date\_understanding} & You are given a known date and asked to calculate a date based on a specific time interval or event. To solve these types of questions, follow these steps: 1. Identify the known date provided in the question. 2. Understand the time interval or event mentioned (e.g., "10 days ago", "one year ago", "5-year anniversary"). 3. Calculate the required date by counting backward or forwards from the known date according to the interval or event. 4. Compare the calculated date with the options given to find the correct answer. Finally, select the option that matches the calculated date. Remember you should output your final answer in the end like <ans>(A)</ans> or <ans>(B)</ans>, etc. \\
\cmidrule{2-2}
~ & Use your knowledge of the calendar and common sense. Carefully read the question and the options. And then, handle the math calculation step by step to find the answer. Finally, select the correct answer.Remember you should output your final answer in the end like <ans>(A)</ans> or <ans>(B)</ans>, etc. \\
\cmidrule{2-2}
~ & Infer the date from context. Remember you should output your final answer in the end like <ans>(A)</ans> or <ans>(B)</ans>, etc. \\
\cmidrule{2-2}
~ &  Given a multi-choice question, your task is to solve the question step by step as follows: 1. Read the question and options. 2. Eliminate the options that are clearly wrong. 3. Find the best option among the remaining ones. Remember you should output your final answer in the end like <ans>(A)</ans> or <ans>(B)</ans>, etc.\\
\midrule
\multirow{4}{*}{movie\_recommendation} & Image you are a movie buff and you are asked to recommend movies similar to a given list of movies. Follow the steps below to find the answer. (1) Identify the common genres and themes of the movies in the list. (2) Consider the time period of production, focusing on classics or movies from a specific era. (3) Compare the options provided, looking for films that match the identified genres, themes, and production period. (4) Select the option that best aligns with the criteria established from the given list of movies. Remember you should output your final answer in the end like <ans>(A)</ans> or <ans>(B)</ans>, etc. \\
\cmidrule{2-2}
~ & Image you are a movie buff and you are asked to recommend movies similar to a given list of movies. You need to think about the genre, the year of production, the actors, and the overall style of the movies. Think step by step and eliminate the options that are not similar to the given list of movies. Choose the option that is most similar to the given list of movies. Remember you should output your final answer in the end like <ans>(A)</ans> or <ans>(B)</ans>, etc. \\
\cmidrule{2-2}
~ & Recommend movies similar to the given list of movies. Remember you should output your final answer in the end like <ans>(A)</ans> or <ans>(B)</ans>, etc. \\
\cmidrule{2-2}
~ & Given a multi-choice question, your task is to solve the question step by step as follows: 1. Read the question and options. 2. Eliminate the options that are clearly wrong. 3. Find the best option among the remaining ones. Remember you should output your final answer in the end like <ans>(A)</ans> or <ans>(B)</ans>, etc. \\
\bottomrule
\\
\caption{Task prompts for different reasoning techniques.}
\label{tab:task-to-prompt}
\end{longtable}
}

\newpage
\section{One-shot In-context Learning Template for Preference Collection}
\label{sec:in-context}
We provide the in-context learning prompt for the student model in the \textbf{Preference Collection} step.
\begin{table}[ht]
    \centering

    \begin{tabular}{p{0.85\linewidth}}
        \toprule
        Task Description: \{Task Description\} \\
        Remember you should include your final answer with the tag \texttt{<ans>} and \texttt{</ans>}. \\
        Question: \{Example Question\} \\
        Answer: Let's think step by step. \{Example Rationale\} \\
        Question: \{Test Question\} \\
        Answer: Let's think step by step. \{to be completed by the student model\} \\
        \bottomrule
        \\
    \end{tabular}    
    \caption{
        One-shot in-context learning prompt for preference collection.
    }
    \label{tab:pref-in-context}
\end{table}

\section{Licenses for existing assets}

The names of the licenses for each asset used in this paper are detailed below.

\begin{table}[ht]
    \centering

    \begin{tabular}{ll}
        \toprule
        \textbf{Asset} & \textbf{License} \\
        \midrule
        BBH & MIT License \\
        GPT-4-LLM & Apache License Version 2.0 \\
        Tulu-v2 & ODC-BY \\
        WizardLM & Apache License Version 2.0 \\
        OpenOrca & MIT License \\
        PIQA & Academic Free License v3.0 \\
        CSQA & MIT License \\
        ARC-E & CC-BY 4.0 \\
        ARC-C & CC-BY 4.0 \\
        GSM8K & MIT License \\
        \midrule
        Gemma-2B & Gemma Terms of Use \\
        Gemma-7B & Gemma Terms of Use \\
        Qwen1.5-1.8B & Tongyi Qianwen RESEARCH LICENSE \\
        CodeGemma-2B & Gemma Terms of Use \\
        Llama-3-70B-Instruction & META LLAMA 3 COMMUNITY LICENSE \\
        MiniCPM-2B & Apache License Version 2.0 \\
        \bottomrule
        \\
    \end{tabular}    
    \caption{
        Licenses for each asset in the paper.
    }
    \label{tab:license}
\end{table}



\clearpage

\end{document}